\def\year{2020}\relax
\documentclass[letterpaper]{article} 
\usepackage{arxiv}  
\usepackage{times}  
\usepackage{helvet} 
\usepackage{courier}  
\usepackage[hyphens]{url}  
\usepackage{amssymb}
\usepackage{amsmath}
\usepackage{multirow}
\usepackage{graphicx} 
\usepackage{graphicx}  
\title{PI-RCNN: An Efficient Multi-sensor 3D Object Detector with Point-based Attentive Cont-conv Fusion Module\thanks{Deng Cai is the corresponding author.}}

\author{Liang Xie\textsuperscript{\rm 1,2}, Chao Xiang\textsuperscript{\rm 1}, Zhengxu Yu\textsuperscript{\rm 1}, Guodong Xu\textsuperscript{\rm 1,2} \\ \textbf{Zheng Yang\textsuperscript{\rm 2}, Deng Cai\textsuperscript{\rm 1,3}, Xiaofei He\textsuperscript{\rm 1,2}} \\ 
\textsuperscript{\rm 1}State Key Lab of CAD\&CG, Zhejiang University, Hangzhou, China\\
\textsuperscript{\rm 2}Fabu Inc., Hangzhou, China\\ 
\textsuperscript{\rm 3}Alibaba-Zhejiang University Joint Institute of Frontier Technologies, Hangzhou, China\\ 
\{lilydedbb, yuzxfred\}@gmail.com  \ \  \{chaoxiang, memoiry, dcai\}@zju.edu.cn \ \ \{yangzheng, xiaofeihe\}@fabu.ai 
}

\begin{document}

\maketitle

\begin{abstract}

LIDAR point clouds and RGB-images are both extremely essential for 3D object detection. So many state-of-the-art 3D detection algorithms dedicate in fusing these two types of data effectively. However, their fusion methods based on Bird’s Eye View (BEV) or voxel format are not accurate. In this paper, we propose a novel fusion approach named Point-based Attentive Cont-conv Fusion(PACF) module, which fuses multi-sensor features directly on 3D points. Except for continuous convolution, we additionally add a Point-Pooling and an Attentive Aggregation to make the fused features more expressive. Moreover, based on the PACF module, we propose a 3D multi-sensor multi-task network called Pointcloud-Image RCNN(PI-RCNN as brief), which handles the image segmentation and 3D object detection tasks. PI-RCNN employs a segmentation sub-network to extract full-resolution semantic feature maps from images and then fuses the multi-sensor features via powerful PACF module. Beneficial from the effectiveness of the PACF module and the expressive semantic features from the segmentation module, PI-RCNN can improve much in 3D object detection. We demonstrate the effectiveness of the PACF module and PI-RCNN on the KITTI 3D Detection benchmark, and our method can achieve state-of-the-art on the metric of 3D AP.

\end{abstract}

\section{Introduction}

\begin{figure}
    \centering
    \includegraphics[width=0.80\linewidth]{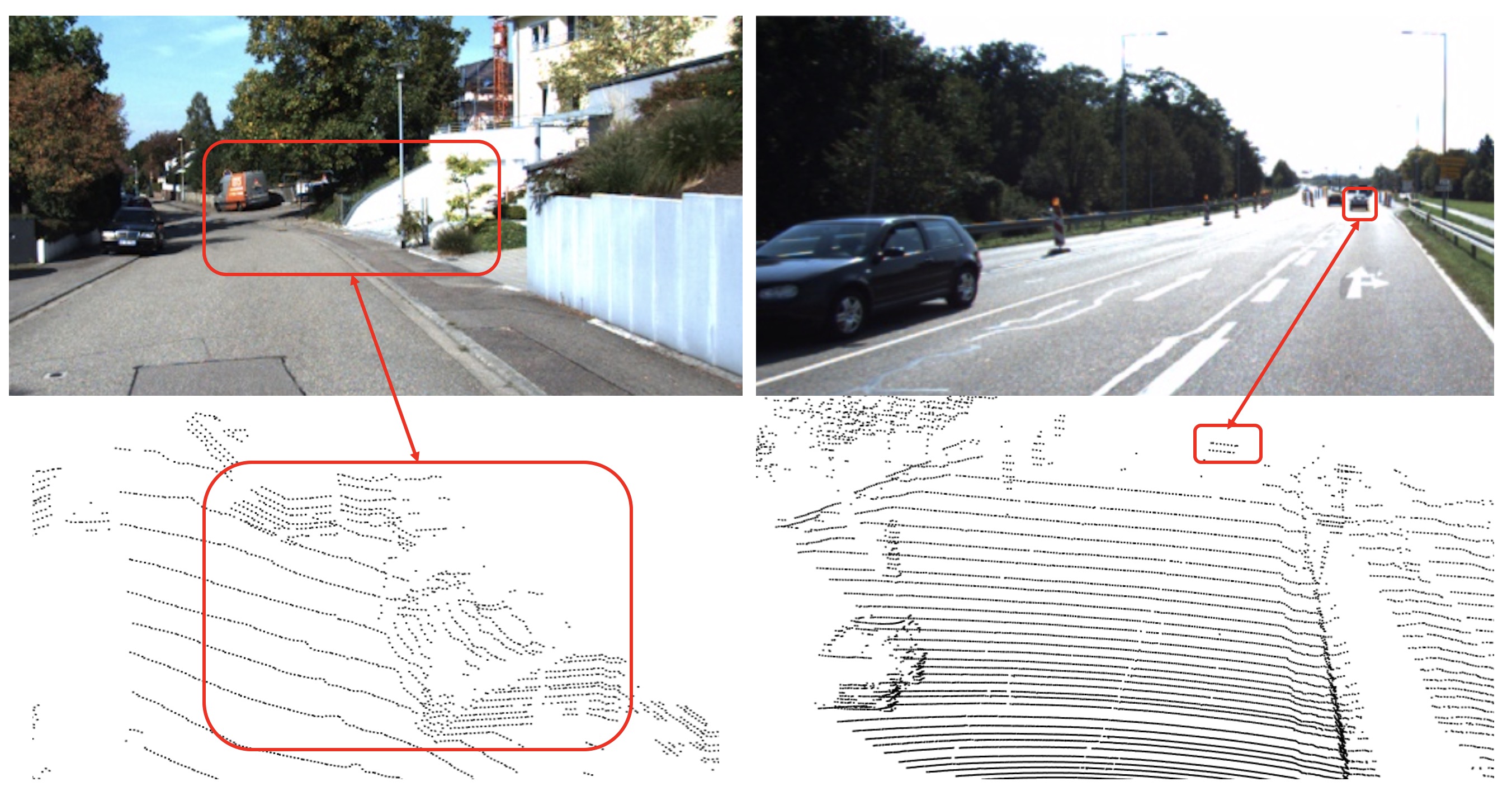}
    \caption{The challenges of LIDAR-based 3D detection. In the left case, we can not distinguish the vehicle from the background only through sparse 3D structure captured by LIDAR. The right case shows that the LIDAR points become too sparse for a car far away, even only several points.}
    \label{figure:Challenges}
\end{figure}

With the rapid development of autonomous driving, 3D detection attracts more and more attention. LIDAR is the most common 3D sensor in autonomous driving. There are existing works detecting 3D objects from LIDAR points\cite{zhou2018voxelnet,yan2018second,yang2018pixor,lang2019pointpillars,shi2019pointrcnn,DBLP:journals/corr/abs-1901-08373}. However, although LIDAR points can capture the 3D structures of objects, they do not have enough semantic information and suffer from the sparsity of points. The loss of semantics causes tough and confusing scenes which the model is hard to tackle. The sparsity of LIDAR points, especially the points far away, brings difficulties for the network to recognize. These challenges are exampled in Figure \ref{figure:Challenges}.

Meanwhile, some works\cite{mousavian20173d,li2019gs3d,ku2019monocular} try to estimate 3D location and dimension of objects via monocular images. Comparing with point clouds, RGB-images have more regular and dense data format and have much richer semantic information to distinguish vehicles and background. However, the nature of 2D image determines that 3D detection algorithms based on monocular images suffer from low precision. 

To address these challenges, many state-of-the-art methods \cite{chen2017multi,ku2018joint,liang2018deep,qi2018frustum,liang2019multi} combine the data of multiple sensors to remedy the semantic loss of point clouds. \cite{chen2017multi,ku2018joint} directly merge the features from images and BEV(birds-eye-view) maps. \cite{qi2018frustum} employ a cascade structure to predict 3D objects via a frustum from the 2D detection bounding box. \cite{liang2018deep} apply continuous convolution\cite{Wang_2018_CVPR} to fuse multi-sensor features.   

However, the direct fusion like \cite{chen2017multi,ku2018joint} ignore the extremely different perspectives of RGB-images and Birds-view maps.
The 3D detection based on frustum\cite{qi2018frustum} suffers from the weakness of 2D detection and involves many points of background or other instances because of occlusion. Although \cite{liang2018deep} apply continuous convolution to overcome the challenge of different perspectives, their fusion based on BEV map is not accurate. BEV-format quantifies the 3D world into a pseudo-image, so the neighbors search and fusion on BEV map suffers from the loss of precision.

To overcome these shortcomings, we propose a novel fusion module named Point-based Attentive Continuous-Convolution Fusion module(PACF module as brief). Different from \cite{liang2018deep,liang2019multi}, we directly apply continuous convolution on raw points. Meanwhile, inspired by some multi-task works\cite{gao2019nddr,liang2019multi}, we combine the image segmentation task and 3D detection to take full advantage of the semantic information from images. Specially, we fuse the semantic features outputted by a segmentation model with the features of LIDAR points via our proposed PACF module. Moreover, based on the PACF module, we propose a robust multi-sensor 3D object detection network named Point-Image RCNN(PI-RCNN as brief).

Our proposed PI-RCNN is inspired by two observations: (1) The most significant information we can obtain from 2D-image is the segmentation mask, and once we obtain the segmentation mask, we naturally get the 2D locations and bounding boxes of objects on images; (2) There is no intersection for objects in 3D space, so we can naturally get the LIDAR points segmentation through only 3D objects label.

PI-RCNN is composed of two sub-networks: an image segmentation sub-network and a point-based 3D detection sub-network. The segmentation sub-network of PI-RCNN is a lightweight fully convolution network, which outputs a prediction mask whose size is the same as the original input image. The detection sub-network is a 3D detector which takes raw LIDAR points as input. The PACF module bridges the two sub-networks and combines the features from RGB-image and LIDAR points to benefit the 3D object detection. With the features fused by our proposed PACF module, our proposed PI-RCNN can effectively improve the performance of 3D object detection. Experiments on KITTI\cite{geiger2013vision} dataset demonstrate the effectiveness of our approach. Our proposed framework PI-RCNN achieves state-of-the-art on the metric of 3D AP.

We summarize our contributions into three aspects:

\begin{quote}
\begin{itemize}

\item We propose a novel fusion method, named PACF module, to fuse the multi-sensor features. PACF module conducts point-wise continuous convolution directly on 3D points and applies a Point-Pooling and an Attentive Aggregation operation to obtain better fusion performance. 
\item Based on the powerful PACF module, we design an efficient multi-sensor 3D object detection algorithm, named Point-Image RCNN(PI-RCNN as brief). What is more, PI-RCNN combines multiple tasks(image segmentation and 3D object detection) to improve the performance of 3D detection.
\item We conduct extensive experiments on KITTI dataset and demonstrate the effectiveness of our approach.
\end{itemize}
\end{quote}

\begin{figure*}[t]
    \centering
    \includegraphics[width=0.70\linewidth]{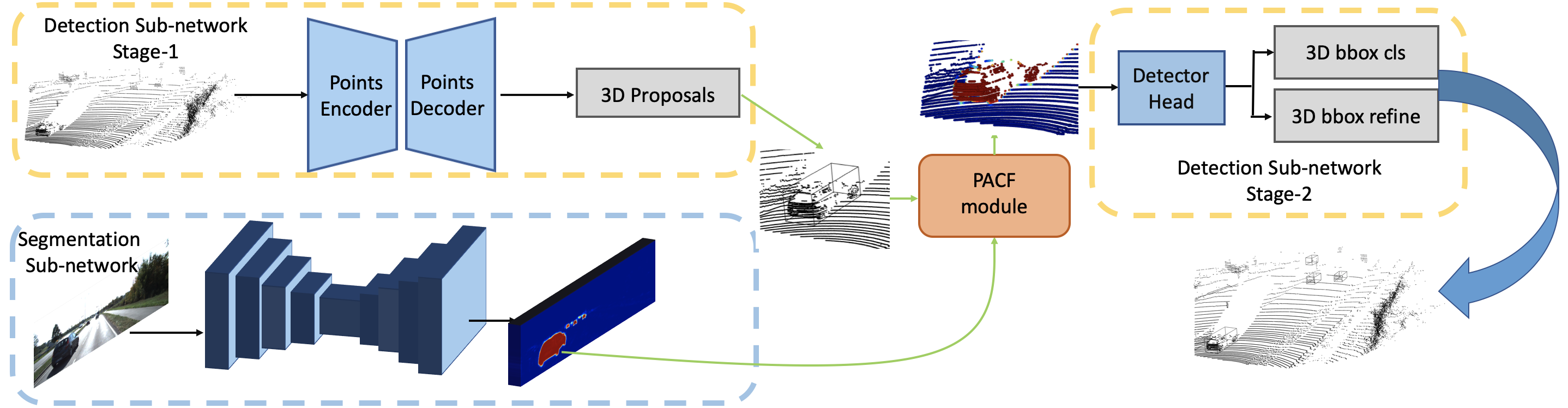}
    \caption{The main architecture of our proposed PI-RCNN. First, an image segmentation sub-network extracts semantic features from RGB-image. Meanwhile, the stage-1 of detection sub-network generate 3D proposals from raw LIDAR points. Then, the 3D points and semantic feature maps are feed into the PACF module to conduct point-wise fusion and supplement the features of points. Finally, the stage-2 of detection sub-network takes the point-wise features augmented from image semantics as input to obtain the final prediction of the 3D bounding box.}
    \label{figure:PI-RCNN}
\end{figure*}

\section{Related Works}

\subsection{3D Object Detection from Single Sensor}

\textbf{3D Object Detection from RGB-images.}  \cite{mousavian20173d,li2019gs3d} employ geometry constrains of 2D bounding box predictions to estimate the pose of 3D objects and obtain the location through camera calibration. \cite{chen20153d} generate 2D proposals from monocular RGB-image and estimate depth map to refine 3D objects' shape and position. \cite{Chen_2016_CVPR} exploit instance and semantic segmentation along with geometric priors to infer 3D object based on monocular images. \cite{wangcvpr2019} generate a set of pseudo points via depth estimation on RGB-image and reason about 3D objects on the generated 3D points. However, due to the lack of depth information, the depth estimation through monocular image is inaccurate, so 3D detection based on RGB-images suffers from low precision.

\textbf{3D Object Detection from Point Clouds.} Due to traditional CNN can not be applied directly on LIDAR points, many algorithms try various ways to address this issue. In the most common paradigm, point clouds are primarily converted to a fixed size pseudo-image which can be processed by a standard CNN, for example, BEV\cite{ku2018joint,liang2018deep,lang2019pointpillars} or voxels\cite{zhou2018voxelnet,yan2018second,yang2018pixor,wang2019voxel}.

There are also algorithms leveraging raw 3D points to detect 3D objects. \cite{qi2017pointnet,qi2017pointnet++} exploit raw points to classify point clouds or predict point segmentation. \cite{shi2019pointrcnn} employ PointNet++\cite{qi2017pointnet++} to generate 3D proposals from raw point clouds and a point-based RCNN to conduct refinement in a local range. 

\subsection{3D Object Detection from Multi Sensors}

\cite{chen2017multi} take RGB-image, front-view, and birds-eye-view as input, and exploits a 3D RPN to generate 3D proposals. \cite{ku2018joint} develop the idea of \cite{chen2017multi}, propose a feature pyramid backbone to extract features from BEV map and merge features from BEV map and RGB-image by a crop and resize operation. \cite{qi2018frustum} use a 3D frustum projected from the 2D bounding box to estimate 3D objects. \cite{liang2018deep} apply continuous convolution\cite{Wang_2018_CVPR} to fuse BEV features with the neighbor points'  features retrieved from the image.

However, the direct fusion methods like \cite{ku2018joint,chen2017multi} are too coarse, the rectangular RoIs(Region of Interest) on images involve lots of background noise and ignore the differences between the perspective of bird's view map and image.  \cite{liang2018deep} employ continuous convolution to avoid the perspective issue, but their BEV-based fusion method suffers the loss of precision, and there is much improvable space to utilize the semantic information of images. Although \cite{liang2019multi} declaim that they apply ``point-wise" continuous convolution, it still conducts fusion on BEV map and does not achieve real ``point-wise" fusion directly on LIDAR points. 

\section{PI-RCNN}

\begin{figure}[t]
    \centering
    \includegraphics[width=0.95\linewidth]{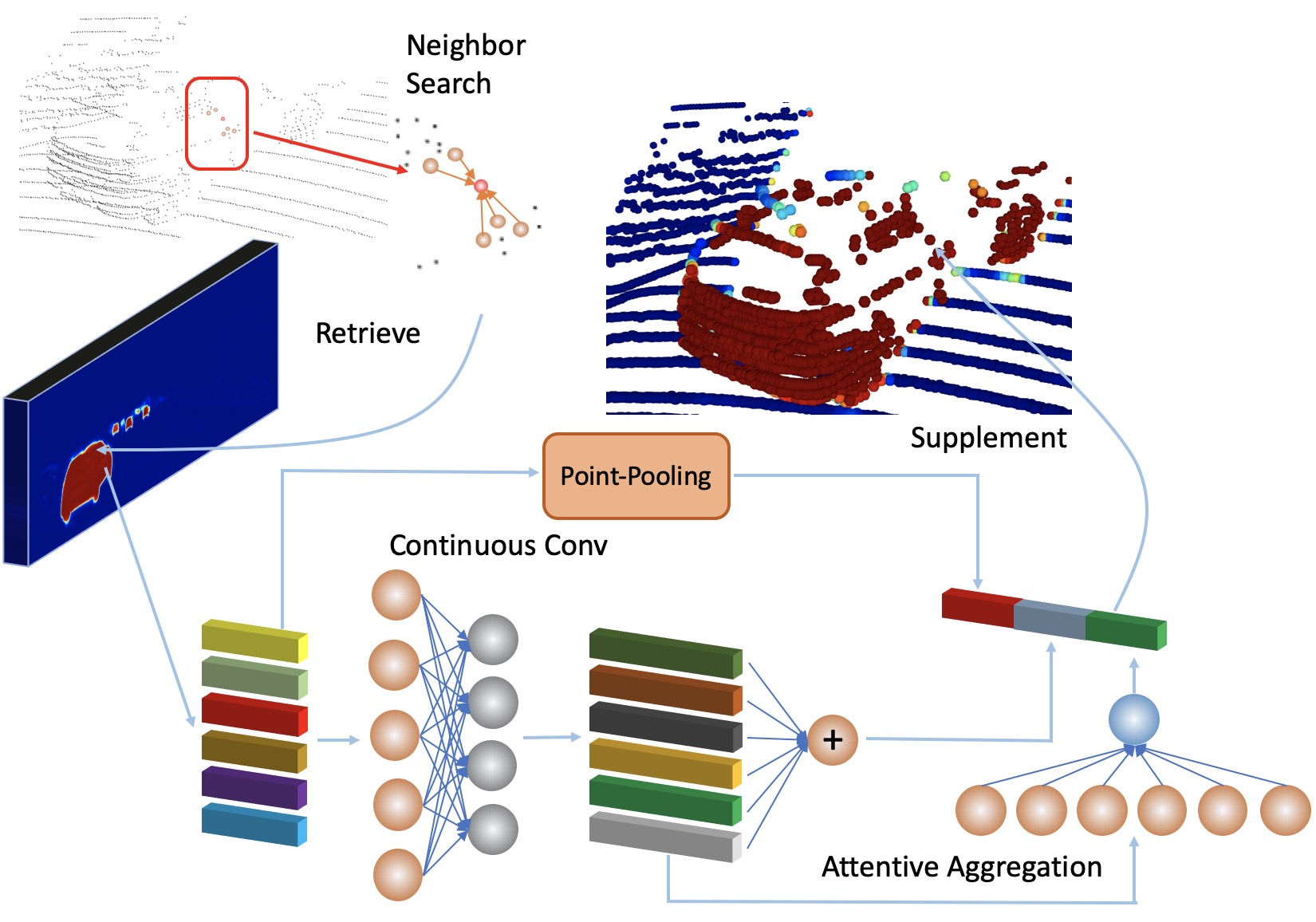}
    \caption{The illustration of our proposed PACF module. PACF module conducts fusion on raw 3D points and retrieves image features from feature maps with more larger resolution and more semantic information. Besides, we add two additional operations: a Point-Pooling along the point-axis to pool the features of neighbor points; an Attentive Aggregation to aggregate features of neighbor through a set of learnable parameters.}
    \label{figure:PACF}
\end{figure}

In this section, we present our proposed novel fusion module, Point-based Attentive Continuous-Convolution Fusion module(PACF module as brief). Different from \cite{liang2018deep,liang2019multi}, PACF module conducts real ``point-wise" continuous convolution directly on 3D LIDAR points and additionally add a Point-Pooling operation and an Attentive Aggregation to make fusion more robust. Moreover, based on the PACF module, we propose Point-Image RCNN (PI-RCNN as brief), a multi-sensor 3D detection network which combines multiple tasks. PI-RCNN combines the image segmentation and 3D object detection and exploits the semantic features from image segmentation to supplement the LIDAR points. The overall architecture of PI-RCNN is illustrated in Figure \ref{figure:PI-RCNN}. PI-RCNN is composed of two sub-networks. One is the segmentation sub-network which takes RGB-images as inputs and outputs semantic features. The other is a point-based 3D detection network, which generates and refines 3D proposals from raw LIDAR points. PACF module is the bridge between the two sub-networks. PACF module conducts fusion operation directly on 3D points instead of BEV or voxel format pseudo-image and merges the semantic features from RGB-image with features from LIDAR points. Moreover, PACF module adds Point-Pooling and Attentive Aggregation to make fused features more expressive. Beneficial from the effectiveness of PACF module, PI-RCNN can detect 3D objects more preciously.

\subsection{Point-based Attentive ContFuse Module}

\textbf{Fusion for multi-sensor data.} The different data format and perspective are the main challenges of fusing features from 2D images and 3D points. RGB-images only represent the 2D projection of the real 3D world on the camera image plane, while LIDAR points capture the 3D structures of the scenes. \cite{chen2017multi,ku2018joint} convert the LIDAR points to BEV(birds-eye-view) pseudo-images and directly fuse the features from BEV maps and RGB-images. However, the proposals on BEV map and RGB-images have different perspectives, so the direct fusion is too coarse to fuse accurate and beneficial features. ContFuse\cite{liang2018deep} project the image features into BEV map and fuse features of the neighbor points with the continuous convolution\cite{Wang_2018_CVPR}. However, BEV-format is only the quantification of the 3D pointclouds and suffers from precious loss, so the neighbor search and fusion on BEV is not accurate, especially in the Z-axis of LIDAR coordinate. Although MMF\cite{liang2019multi} build a dense correspondence between image and BEV, they still do not apply real ``point-wise" continuous convolution directly on 3D points. 

\textbf{PACF module.} To address these issues, we propose a novel fusion module, PACF module, which achieves more accurate and robust fusion. The details of the PACF module are illustrated in Figure \ref{figure:PACF}. Given a feature map extracted from RGB-image and raw LIDAR points, PACF module outputs a set of discrete 3D points whose features contains the semantic information from RGB-image. In detail, the PACF module consists of five steps. (1) We search the $k$ nearest neighbor points in a distance range $d$ ($d = +\infty$ as default) for each 3D point. (2) We project the neighbor points onto the feature maps extracted from the 2D image plane via camera calibration. (3) We retrieve the corresponding semantic features from images and combine image features with the geometric offset of 3D points. (4) We exploit attentive continuous convolution to fuse the semantic+geometric features of k-nearest neighbor points. (5) We conduct a Point-Pooling operation for the outputs of step (3) and concatenate them with outputs of step (4) as the final features of target points.

The attentive continuous convolution is improved based on ContFuse\cite{liang2018deep}. We denote $x_i$ as the coordinate of point $p_i$, $f_i$ as the concatenation of point features outputted by detection sub-network and the semantic features retrieved from the output of segmentation sub-network. Note, we concatenate the  semantic features and point features outputted by detection sub-network the fused features, so $f_i$ is a $(C_{seg} + C_{lidar})$-d vector, where $C_{seg}$ is the channel number of the semantic features and $C_{lidar}$ is the channel number of the point features. The continuous convolution is defined as:

\begin{equation}
y_{cc, k}^{i} = \textsc{MLP}_{cc}(f'_k), \ \ f'_{k} = \textsc{concat}(f_k, x_k - x_i)
\label{equation_1}
\end{equation}

\begin{equation}
y_{cc}^{i} = \sum_{k}{y_{cc, k}^{i}}
\end{equation}

\noindent where $i = 1, 2, ..., N$ and N is the number of LIDAR points, $k = 1, 2, ..., K$ and K is the number of neighbor points (including ego point), $x_i$ is the coordinate of target point $p_i$, $x_k$ is the coordinate of neighbor points $p_k \in \text{Neighbor}(p_i)$, so $x_k - x_i$ represents the geometric offset from the target point $p_i$ to the neighbor point $p_k$, $y_{cc, k}^{i}$ is a $D_o$-d row vector, and $y_{cc}^{i}$ is the output of continuous convolution. $\textsc{MLP}_{cc}$ in Equation \ref{equation_1} approximates continuous convolution, which converts the $K \times D_i$ input into the $K \times D_o$ output, where $D_i=C_{seg}+C_{lidar}+3, D_o$ are channel numbers of the input and output features respectively. 

Inspired by the Pooling operation in CNN and attentive mechanism, we add a Point-Pooling operation and an Attentive Aggregation to strengthen the continuous convolution. In detail, we conduct a Pooling operation on the features of $K$ neighbor points. The Point-Pooling can be represented as:

\begin{equation}
y_{pool}^{i} = \textsc{POOL}(F'), \ \ F' = [f'^T_{1}, f'^T_{2}, ..., f'^T_{K}]^T
\end{equation}

\noindent where $F' \in \mathcal{R}^{K \times D_i}$ is the features of all neighbors, $y_{pool}^{i}$ represents the pooled features for each target point $i$. The $\textsc{POOL}$ is conducted along the point-axis. In practice, we exploit Max-Pooling to obtain the most expressive features from $K$ neighbor points. Besides, we conduct an Attentive Aggregation to merge the features of $K$ neighbor points. In practice, we employ another MLP to aggregate neighbors, that is to say, for each target point $i$:

\begin{equation}
y_{a}^{i} = \textsc{MLP}_{aggr}(Y_{cc}^{i}) = \sum_{k} w_{k} y_{cc,k}^{i}
\end{equation}

\noindent where $Y_{cc}^{i} \in \mathcal{R}^{K \times D_o}$ represents the features of $K$ neighbor points outputted by the $ \textsc{MLP}_{cc}$, the $\textsc{MLP}_{aggr}$ aggregates the $K \times D_o$ neighbor features into $D_o$-d features of target point through a set of learnable parameters. The final output of the PACF module is the concatenation of above three parts:

\begin{equation}
y_{o}^{i} = \textsc{concat}(y_{cc}^{i}, y_{a}^{i}, y_{pool}^{i})
\end{equation}

\textbf{Improvements comparing with previous methods.} Our proposed PACF module has five differences from \cite{liang2018deep,liang2019multi}. Primarily, they both fuse features on the pixels of BEV. However, BEV format quantifies the real 3D space to a 2D pseudo-image, so the neighbor search and feature fusion applied on the pixels on BEV is not accurate. In contrast, we conduct the neighbor search, continuous convolution, and final fusion directly on raw 3D points instead of BEV, which precludes the quantification loss. Secondly, except for the MLP for continuous convolution, we add another learnable MLP to fuse the features from neighbor points, which can be considered as an attention mechanism for the features of neighbors. Thirdly, to avoid the interpolation loss, we retrieve the image features on features map with a larger resolution, whose size is consistent with the original size of the image. The fourth difference is that we combine the image segmentation task with 3D object detection. Instead of using the image features learned from 3D detection task, we first pre-train the image sub-network on a segmentation dataset. In the Experiments Section, we conduct experiments to compare the features pre-trained on segmentation task with the features learned from 3D detection. We argue that the features learned under the supervision of semantic segmentation are more expressive, and the combination of multiple tasks (image segmentation and 3D detection) is robust. Finally, inspired by the pooling operation in CNN and the attentive mechanism, we conduct point-wise pooling among the features of neighbor points and add a learnable Attentive Aggregation operation to merge the features of neighbors more effectively.

We argue that these improvements make a significant difference. In the Experiments Section, we will conduct ablation experiments to analyze the effects of these differences.

\subsection{Main Architecture of PI-RCNN}

PI-RCNN is a multi-task 3D detection network and is composed of two sub-networks: image segmentation sub-network and 3D Detection sub-network.

\textbf{Semantic Segmentation Sub-Network.} To obtain robust semantic features from RGB-images, we first analyze which features from images are most beneficial for 3D objects detection. For the 2D object detection task, the feature extractor is usually pre-trained on classification dataset, such as ImageNet\cite{deng2009imagenet}, which is sufficient enough for detecting 2D bounding box. Because the target of 2D object detection is only predicting the rectangular bounding box, which does not demand meticulous features in 2D proposals. As long as the features of RoI capture the part region of objects, the detector's head can classify and regress the proposals correctly. However, it is insufficient for the dense correspondence between image pixels and LIDAR points.

\begin{figure}[t]
    \centering
    \includegraphics[width=0.90\linewidth]{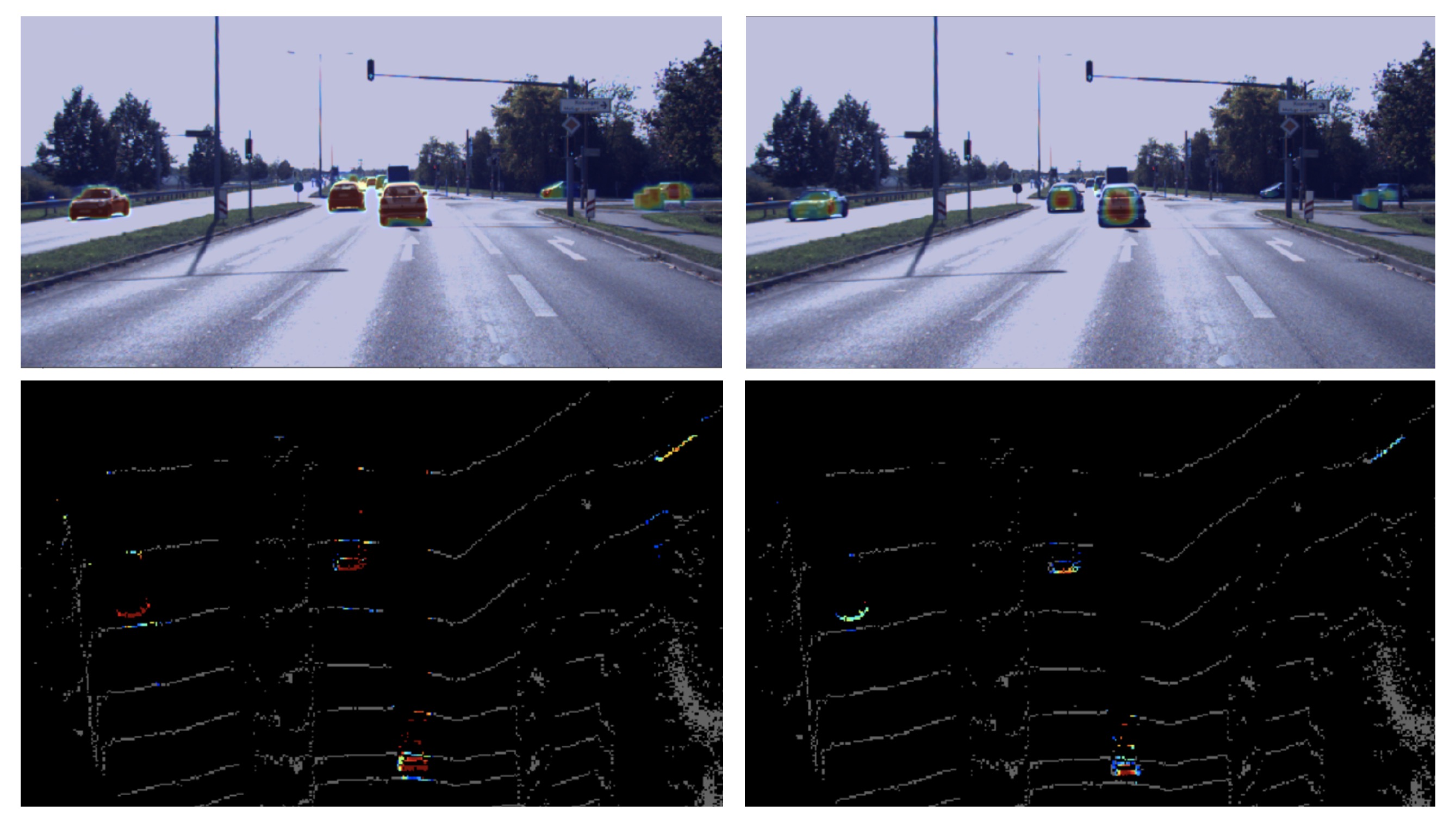}
    \caption{The top images are the segmentation prediction outputted by segmentation sub-network. The bottom images are the LIDAR points in the Birds-eye-view, and the color of points is corresponding with the values retrieved from the segmentation mask. Left is the case of pretraining on the segmentation dataset, while the right is the case of end-to-end training only under the supervision of 3D detection label.}
    \label{figure:seg_pretrain}
\end{figure}

We argue that image features learned from 3D detection label are too coarse for the correspondence between image pixels and 3D points. We observed that once we get the segmentation mask from RGB-image, we can project the 3D points onto the 2D image plane to retrieve the corresponding segmentation of 3D points. Because segmentation mask is a pixel-level prediction, which does not involve the background pixels like the 2D bounding box, it can give each point more accurate semantic information to help the detection sub-network to predict 3D objects more preciously. The comparison of features outputted by pre-trained segmentation sub-network and no pre-training sub-network is illustrated in Figure \ref{figure:seg_pretrain}. Therefore, we combine the image segmentation task with 3D detection and use the outputs from a segmentation network as the semantic features of RGB-images. Besides, segmentation feature maps have a larger resolution than the outputs of classification backbone, which makes the projection and fusion between LIDAR points and image pixels more accurate. In the Experiments Section, we will conduct a relative ablation study to verify the effectiveness of pre-training on segmentation dataset. Note that we do not need pre-train an instance-level segmentation sub-networks, because the target of segmentation supervision is only helping us obtain semantic features for fusion and we detect objects based on LIDAR points. We exploit UNet\cite{ronneberger2015u}, a lightweight fully-convolution network, as the segmentation sub-network of PI-RCNN. Note, in practice, we can alternate it with other lightweight segmentation networks. 

\textbf{3D Detection Sub-Network} We argue that point-wise fusion is more robust than fusion based on BEV map. To conduct the point-wise fusion operation, we need to employ a 3D detection network based on raw 3D points. Therefore, we employ PointRCNN\cite{shi2019pointrcnn}, a two-stage 3D detection network whose inputs are raw LIDAR points, as the detection Sub-Network of PI-RCNN. PointRCNN employ PointNet++\cite{qi2017pointnet++} as its first stage to generate 3D proposals from raw LIDAR points. Its stage-2 transforms the points in each proposal to canonical coordinates to refine the 3D bounding box. 

\subsection{Fusion Strategy}

\begin{figure}[t]
    \centering
    \includegraphics[width=0.95\linewidth]{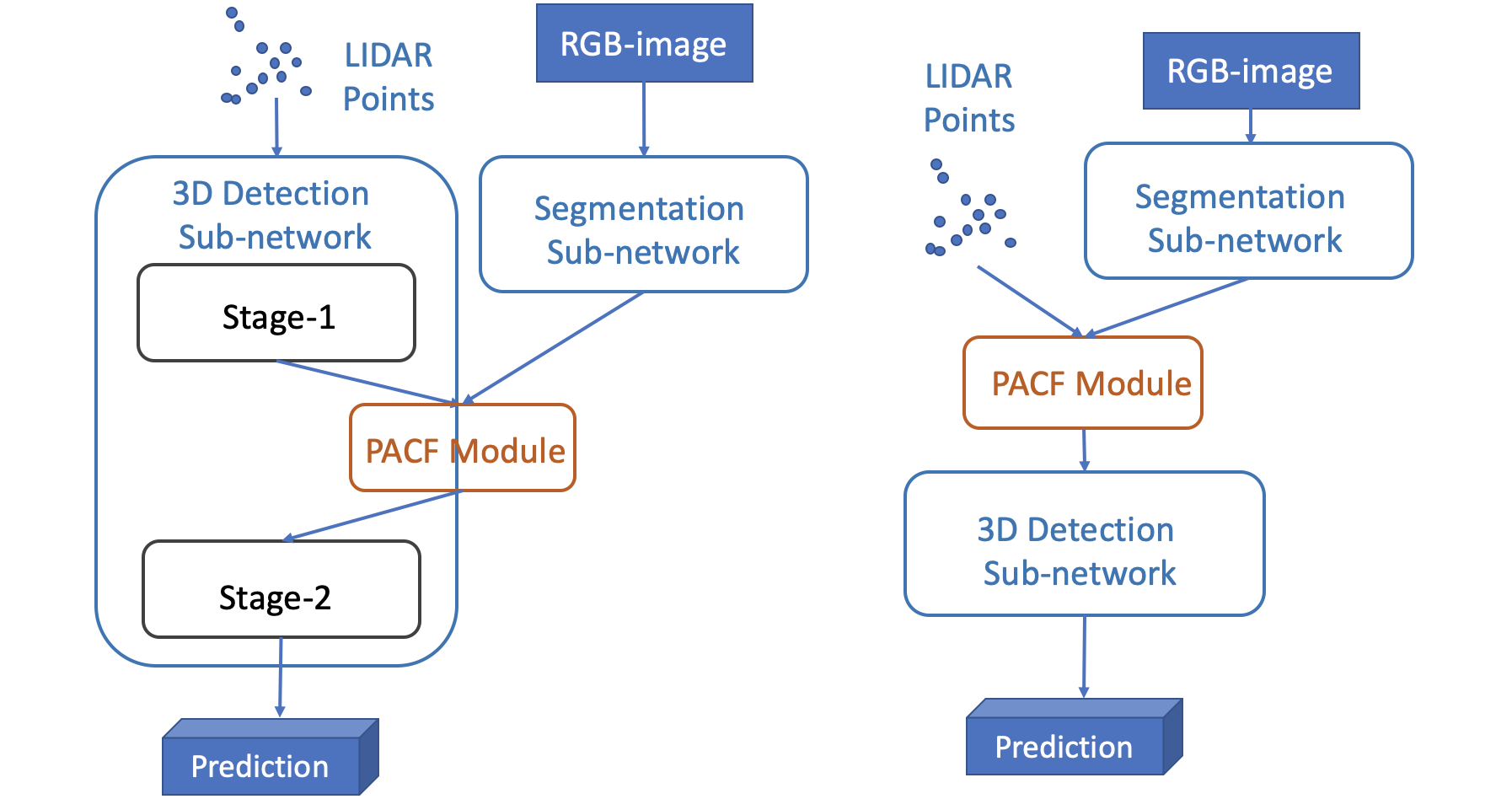}
    \caption{PI-RCNN V1(left) and V2(right).}
    \label{figure:V1V2}
\end{figure}

We provide two fusion strategies. The main difference between the two strategies is the location of the fusion module. The comparison is illustrated in Figure \ref{figure:V1V2}. We denote these two versions of fusion strategy as \textbf{PI-RCNN V1} and \textbf{PI-RCNN V2} respectively. In the Experiments Section, we will analyze the performance of two fusion strategies.

\textbf{PI-RCNN V1}: We fuse the features from multiple sensors in the ``middle-way", as illustrated in Figure \ref{figure:V1V2}. In this strategy, the semantic features from image act as a supplementation of the 3D points features outputted by the first stage of detection sub-network. 

\textbf{PI-RCNN V2}: We can also conduct the fusion operation at the beginning of the detection network. After obtaining the output of segmentation sub-network, we concatenate the image features with raw LIDAR points as the input of detection sub-network. For this fusion strategy, we can alternate the detection sub-network with other 3D detectors which takes inputs of arbitrary format. For example, when leveraging a 3D detection algorithm based on the format of the BEV map or voxels, the semantic features can act as the extra features of LIDAR points.

\subsection{Loss}

For the 3D detection sub-network, we follow the loss function introduced by the \cite{shi2019pointrcnn}. The loss of detection sub-network is defined as:

\begin{equation}
\mathcal{L}_{\text{det}} = \mathcal{L}_{\text{reg}} + \mathcal{L}_{\text{refine}}
\end{equation}

\noindent where $\mathcal{L}_{\text{reg}}, \mathcal{L}_{\text{refine}}$ are defined the same as original paper.

For the training of image segmentation sub-network, we need a semantic segmentation label as supervision. As mentioned in \cite{shi2019pointrcnn}, the 3D objects are not overlapped with each other, and we can get the segmentation of points from the 3D detection label. Hence, we can obtain a sparse segmentation mask by projection the points segmentation onto the 2D image plane, and we only compute loss on the pixels with supervision. To address the imbalance between the foreground and background, we employ Focal Loss\cite{lin2017focal} as:

\begin{equation}
\mathcal{L}_{\text{seg}}(p_t) = - \alpha_{t}(1 - p_t)^{\gamma} \log(p_t)
\end{equation}

\noindent where  $p_t = p$ for forground point otherwise $1 - p$, $p$ is the scores outputted by network. And we keep the default settings $\alpha_t = 0.25, \gamma = 2$ as the original paper. 

Therefore, the total loss is:

\begin{equation}
\mathcal{L} = \mathcal{L}_{\text{det}} + \lambda \mathcal{L}_{\text{seg}}
\end{equation}

\noindent where $\lambda$ is the weight of segmentation loss. For the sake of simplicity, we use $\lambda = 1$ as the default setting.

Although our proposed PI-RCNN can be trained end-to-end without pretraining on segmentation dataset, we observe that initialization is essential for the performance of 3D detection. So in practice, we pre-train the segmentation sub-network on a semantic segmentation dataset and fix the parameters of the segmentation sub-network when training the detection sub-network. 

\section{Experiments}

\subsection{Implementation and Training Details}

\textbf{Network Architecture.} For the segmentation sub-network, considering the need for real-time detection, we follow the network structure of UNet\cite{ronneberger2015u}, a lightweight and fully-convolution network. The segmentation sub-network can be alternated with other segmentation networks. Because our primary goal is using semantic features to improve the performance of 3D object detection, so we do not pay much attention to the architecture of segmentation sub-network and employ the same settings for the segmentation sub-network for all experiments.

For the 3D detection sub-network, we exploit a point-based 3D detection algorithm, PointRCNN\cite{shi2019pointrcnn}. PointRCNN is a two-stage 3D detector and predicts 3D objects directly by raw LIDAR points. To compare fairly, in all experiments, we use consistent settings with the original paper. Note, if we use the ``PI-RCNN V2" fusion strategy, theoretically, we can alternate the detection sub-network with almost any other 3D detection algorithm based on LIDAR points, whatever format of input it takes. For the sake of simplicity, in all the following experiments, we employ the ``PI-RCNN V1" fusion strategy as default.

\textbf{Input Representation.} For the detection sub-network, we take raw 3D points as the input, instead of BEV or voxel format. We follow the settings in \cite{shi2019pointrcnn} for the 3D points input. We set the region of concern of LIDAR points as $[0, 70.4] \times [-40, 40] \times [-1, 3]$ in LIDAR coordinate and subsample 16,384 points in the viewable region of camera as inputs. For the RGB-image, we resize the RGB-image to $376 \times 1248$ due to the demand of upsampling operation in the segmentation sub-network. When testing, we find that sampling the input points like training is better than inputting all the points. So we test all our models with the same subsampling strategy. Although this will bring some randomness to the evaluation results, we find that the results are stable($\pm 0.10$ for 3D AP(M)) for one model.

\textbf{Data Augmentation.} To guarantee the correct correspondence between LIDAR points and image pixels, we do not use GT-AUG mentioned in PointRCNN when training. This is different from most 3D detection algorithms based only on LIDAR.

When pretraining the segmentation sub-network, we apply data augmentation to obtain better performance. In detail, we randomly flip the image horizontally, and randomly center-crop the image with a ratio 0.8. Besides the spatial augmentation, we enhance the brightness, contrast, saturation of images with a random factor in $[0.9, 1.1]$. We apply all above augmentations with a probability 0.5.

\subsection{Results on KITTI Dataset}

\begin{table}[t]
	\small 
	\begin{center}
		\scalebox{0.975}{
\scalebox{0.9}{
			\begin{tabular}{c||ccc}
				\hline
				\multirow{2}*{Method} & 			
				\multicolumn{3}{c}{3D AP}\\
				&Easy & Moderate & Hard \\
				\hline\hline
				VoxelNet \cite{zhou2018voxelnet}* & 77.47 & 65.11 & 57.73  \\
				PointPillar \cite{lang2019pointpillars}* & 79.05 & 74.99 & 68.30  \\
				MV3D \cite{chen2017multi}+ & 71.09 & 62.35 & 55.12  \\
				ContFuse \cite{liang2018deep}+ & 82.54 & 66.22 & 64.04  \\
				AVOD-FPN \cite{ku2018joint}+ & 81.94 & 71.88 & 66.38  \\
				F-PointNet \cite{qi2018frustum}+ & 81.20 & 70.39 & 62.19  \\
				\hline 
				PointRCNN \cite{shi2019pointrcnn}* & 83.25 & 74.59 & 70.01 \\
				PI-RCNN(Ours)+ & \textbf{84.37} & \textbf{74.82} & \textbf{70.03} \\
				\hline
			\end{tabular}
}
		}	
	\end{center}
	\caption{Performance comparison of 3D AP(Average Precision) with previous methods on KITTI \textit{testing} split. The methods followed by ``*" take only LIDAR points as input, while methods followed by ``+" use both LIDAR points and RGB-images. The results of PointRCNN are based on our re-implementation without GT-AUG.}
	\label{tab:test_performance}
	\vspace{-3mm}
\end{table}

\begin{table}[t]
	\small 
	\begin{center}
		\scalebox{0.975}{
		
\scalebox{0.9}{
			\begin{tabular}{c|ccc}
				\hline
				\multirow{2}*{Method} & \multicolumn{3}{c}{3D AP(Car)} \\
				&Easy & Moderate & Hard \\
				\hline\hline
				MV3D \cite{chen2017multi} & 71.29 & 62.68 & 56.56 \\
				ContFuse \cite{liang2018deep} & 82.54 & 66.22 & 64.04 \\
				AVOD-FPN \cite{ku2018joint} & 84.41 & 74.44 & 68.65 \\
				F-PointNet \cite{qi2018frustum} & 83.76 & 70.92 & 63.65 \\
				\hline
				\footnotesize PointRCNN \cite{shi2019pointrcnn} & 86.42 & 77.10 & 76.11 \\
				PI-RCNN & \textbf{88.27} & \textbf{78.53} & \textbf{77.75} \\
				\hline
			\end{tabular}

}
		}	
	\end{center}
	\caption{Performance comparison of 3D AP with previous methods on KITTI \textit{val} split. The results of PointRCNN are based on our re-implementation without GT-AUG.}
	\label{tab:val_performance}
	\vspace{-3mm}
\end{table}

\begin{table}[t]
	\small 
	\begin{center}
		\scalebox{0.975}{
			\begin{tabular}{c|cc|ccc}
				\hline
				\multirow{2}*{PACF} & \multirow{2}*{PointPool} & \multirow{2}*{Att Aggr} & \multicolumn{3}{c}{3D AP(Car)} \\
				&&&Easy & Moderate & Hard \\
				\hline\hline
				No & - & - & 86.42 & 77.10 & 76.11 \\
			    Yes & $\times$  & $\times$  & 87.77 & 77.96 & 76.92 \\
			    Yes & \checkmark & $\times$  & 88.23 & 78.42 & 77.23 \\
			    Yes & $\times$ & \checkmark & 87.98 & 78.22 & 76.97 \\
			    Yes & \checkmark & \checkmark & \textbf{88.27} & \textbf{78.53} & \textbf{77.75} \\
				\hline
			\end{tabular}
		}	
	\end{center}
	\caption{Ablation study about the effects of Point-Pooling and Attentive Aggregation operation. \textit{No PACF} represents the baseline of our re-implemented PointRCNN. }
	\label{tab:ablation_PACF}
	\vspace{-3mm}
\end{table}

\begin{table}[t]
	\small 
	\begin{center}
		\scalebox{0.975}{
			\begin{tabular}{c|ccc}
				\hline
				\multirow{2}*{K} & \multicolumn{3}{c}{3D AP(Car)} \\
				& Easy & Moderate & Hard \\
				\hline\hline
				1 & 87.31 & 77.59 & 75.88 \\
				3 & \textbf{88.27} & \textbf{78.53} & \textbf{77.75} \\
				5 & 87.34 & 77.98 & 77.38 \\
				10 & 86.33 & 77.15 & 75.27 \\
				\hline
			\end{tabular}
		}	
	\end{center}
	\caption{Ablation study about the K.}
	\label{tab:ablation_PACF_K}
	\vspace{-3mm}
\end{table}

\begin{table}[t]
	\small 
	\begin{center}
		\scalebox{0.975}{
			\begin{tabular}{c|ccc}
				\hline
				\multirow{2}*{Method} & \multicolumn{3}{c}{3D AP(Car)} \\
				&Easy & Moderate & Hard \\
				\hline\hline
				PI-RCNN V1 & \textbf{88.27} & \textbf{78.53} & \textbf{77.75} \\
				PI-RCNN V2 & 87.66 & 78.01 & 76.55 \\
				\hline
			\end{tabular}
		}	
	\end{center}
	\caption{The performance comparison of V1 and V2.}
	\label{tab:ablation_v1v2}
	\vspace{-3mm}
\end{table}

\begin{table}[t]
	\small 
	\begin{center}
		\scalebox{0.975}{
			\begin{tabular}{c|ccc}
				\hline
				\multirow{2}*{Image Features} & \multicolumn{3}{c}{3D AP(Car)} \\
				&Easy & Moderate & Hard \\
				\hline\hline
				single class & 86.23 & 77.16 & 76.16 \\
				multi classes & \textbf{88.27} & \textbf{78.53} & \textbf{77.75} \\
				\hline
			\end{tabular}
		}	
	\end{center}
	\caption{Ablation study about the image features.}
	\label{tab:ablation_image_features}
	\vspace{-3mm}
\end{table}

\begin{table}[!htbp]
	\small 
	\begin{center}
		\scalebox{0.975}{
			\begin{tabular}{c|ccc}
				\hline
				\multirow{2}*{Pre-train} & \multicolumn{3}{c}{3D AP(Car)} \\
				&Easy & Moderate & Hard \\
				\hline\hline
				No & 85.98 & 76.34 & 74.88 \\
				Yes & \textbf{86.23} & \textbf{77.16} & \textbf{76.16} \\
				\hline
			\end{tabular}
		}	
	\end{center}
	\caption{Ablation study about the pre-training of the segmentation sub-network.}
	\label{tab:ablation_pretrain}
	\vspace{-3mm}
\end{table}

We evaluate PI-RCNN on KITTI \cite{geiger2013vision} dataset. KITTI 3D detection dataset contains 7481 \textit{training} samples and 7518 \textit{testing} samples. The training samples are provided with labels, while the results in \textit{testing} set must be submitted to the official test server to evaluate. We follow the common \textit{train/val} split mentioned in \cite{chen2017multi} to divide 7481 \textit{training} samples into \textit{train} split with 3712 samples and \textit{val} split with 3769 samples. We evaluate our approach on \textbf{Car} class and compare PI-RCNN with state-of-the-art 3D detectors on both \textit{val} split and \textit{testing} split of KITTI dataset. For all the following experiments, the models are trained on \textit{train} split and evaluated on \textit{val} or \textit{test} split.

We compare PI-RCNN with other state-of-the-art methods on both \textit{testing} and \textit{val} split. The evaluation results on \textit{testing} and \textit{val} set are shown in Table \ref{tab:test_performance} and \ref{tab:val_performance} respectively. We follow the implementation released by PointRCNN\cite{shi2019pointrcnn}. Note, we do not use the GT-AUG mentioned in PointRCNN when training on 3D detection task due to the need of multi-sensor fusion. So when comparing with PointRCNN, we only compare with the results of our re-implementation without GT-AUG on the \textit{testing/val} split. On the \textit{testing} split, PI-RCNN surpasses the previous state-of-the-art methods on the metric of 3D AP. On the \textit{val} split, PI-RCNN outperforms the state-of-the-art multi-sensor 3D detectors. Meanwhile, our PI-RCNN outperforms the baseline PointRCNN on both \textit{testing} and \textit{val} in the absence of GT-AUG. The results demonstrate the effectiveness of our proposed PI-RCNN.

\subsection{Ablation Study}

We conduct ablation studies to analyze the effects of the PACF module and PI-RCNN. All models are trained on the \textit{train} split and evaluated on the \textit{val} split of KITTI dataset. All evaluations on the \textit{val} split are performed via 40 recall positions instead of the 11 recall positions.

\textbf{PACF module.} We conduct some ablation experiments about PACF module. We first analyze the effect of hyper-parameter $K$, and the results are shown in Table \ref{tab:ablation_PACF_K}. As mentioned in \cite{liang2018deep}, the continuous convolution might learn to ignore the noise of distant points, so for the sake of simplicity, we use $d=+\infty$ for all experiments. The best result comes from the $K=3$ setting. Meanwhile, we observe that $K=5$ and $K=10$ are even worse than $K=1$. The reason might be that large $K$ involves distant points and brings noises for the features of the target point. Then we study the effects of the Point-Pooling and Attentive Aggregation. The results are shown in Table \ref{tab:ablation_PACF}. Table \ref{tab:ablation_PACF} shows that the additional Point-Pooling and Attentive Aggregation operations are beneficial for the feature fusion.

\textbf{PI-RCNN V1 vs. V2.} As mentioned above, there are two fusion strategies we can choose. We analyze these two versions of PI-RCNN. The comparison results are shown in Table \ref{tab:ablation_v1v2}. We can see that V1 slightly outperforms V2 and the results suggest that fusion in the ``middle" of detection sub-network is better than fusion at the beginning. One possible reason might be that the stage-1 of detection sub-network learns to generate 3D proposal mainly through the 3D information of LIDAR points and the supplementary features appended at the beginning do not contribute as much as fusion in the "middle".

\textbf{Semantic Features.} Fusing image features of which layer in the segmentation sub-network is important for PI-RCNN performance. Table \ref{tab:ablation_image_features} shows the effects of different image features, where \textit{single class} represents that the category we interest is the only foreground class when training seg sub-network. \textit{multi classes} represents training the seg sub-entwork with all categories. Table \ref{tab:ablation_image_features} shows that \textit{multi classes} setting gets the best results. The reason might be that the network might comprehend the whole scene more preciously if we can give it the priors of more categories. For example, if the network could know some points belong to the buildings or other background, these points would be wrongly recognized less possibly.

\textbf{Segmentation Pretraining.} As mentioned above, our multi-task model PI-RCNN can be trained end-to-end only under the supervision of 3D objects annotation. However, the results in Table \ref{tab:ablation_pretrain} suggest that pre-training the segmentation sub-network improves performance. Note that the segmentation sub-network of \textit{pretrain} model are trained with \textit{single class} training strategy, because the 3D objects annotation only provide the supervision for binary classes(a foreground \textit{Car} class we interest and background classes).

\section{Conclusion}

In this paper, we propose a Point-based Attentive Cont-Conv Fusion(PACF) module and a multi-sensor multi-task 3D object detection network named PI-RCNN. PI-RCNN combines the image segmentation and 3D detection. Our proposed framework is simple but effective. Our proposed PI-RCNN achieves the state-of-the-art results on KITTI 3D Detection benchmark.

\section*{Acknowledgments}

This work was supported in part by The National Key Research and Development Program of China (Grant Nos: 2018AAA0101400), in part by The National Nature Science Foundation of China (Grant Nos: 61936006).

\clearpage

\bibliographystyle{aaai}
\bibliography{reference.bib}

\begin{thebibliography}{}

\bibitem[\protect\citeauthoryear{Chen \bgroup et al\mbox.\egroup
  }{2015}]{chen20153d}
Chen, X.; Kundu, K.; Zhu, Y.; Berneshawi, A.~G.; Ma, H.; Fidler, S.; and
  Urtasun, R.
\newblock 2015.
\newblock 3d object proposals for accurate object class detection.
\newblock In {\em Advances in Neural Information Processing Systems},
  424--432.

\bibitem[\protect\citeauthoryear{Chen \bgroup et al\mbox.\egroup
  }{2016}]{Chen_2016_CVPR}
Chen, X.; Kundu, K.; Zhang, Z.; Ma, H.; Fidler, S.; and Urtasun, R.
\newblock 2016.
\newblock Monocular 3d object detection for autonomous driving.
\newblock In {\em The IEEE Conference on Computer Vision and Pattern
  Recognition (CVPR)}.

\bibitem[\protect\citeauthoryear{Chen \bgroup et al\mbox.\egroup
  }{2017}]{chen2017multi}
Chen, X.; Ma, H.; Wan, J.; Li, B.; and Xia, T.
\newblock 2017.
\newblock Multi-view 3d object detection network for autonomous driving.
\newblock In {\em Proceedings of the IEEE Conference on Computer Vision and
  Pattern Recognition},  1907--1915.

\bibitem[\protect\citeauthoryear{Deng \bgroup et al\mbox.\egroup
  }{2009}]{deng2009imagenet}
Deng, J.; Dong, W.; Socher, R.; Li, L.-J.; Li, K.; and Fei-Fei, L.
\newblock 2009.
\newblock Imagenet: A large-scale hierarchical image database.
\newblock In {\em 2009 IEEE conference on computer vision and pattern
  recognition},  248--255.
\newblock Ieee.

\bibitem[\protect\citeauthoryear{Gao \bgroup et al\mbox.\egroup
  }{2019}]{gao2019nddr}
Gao, Y.; Ma, J.; Zhao, M.; Liu, W.; and Yuille, A.~L.
\newblock 2019.
\newblock {NDDR}-{CNN}: Layerwise feature fusing in multi-task cnns by neural
  discriminative dimensionality reduction.
\newblock In {\em IEEE International Conference on Computer Vision and Pattern
  Recognition (CVPR)}.

\bibitem[\protect\citeauthoryear{Geiger \bgroup et al\mbox.\egroup
  }{2013}]{geiger2013vision}
Geiger, A.; Lenz, P.; Stiller, C.; and Urtasun, R.
\newblock 2013.
\newblock Vision meets robotics: The kitti dataset.
\newblock {\em The International Journal of Robotics Research}
  32(11):1231--1237.

\bibitem[\protect\citeauthoryear{Ku \bgroup et al\mbox.\egroup
  }{2018}]{ku2018joint}
Ku, J.; Mozifian, M.; Lee, J.; Harakeh, A.; and Waslander, S.~L.
\newblock 2018.
\newblock Joint 3d proposal generation and object detection from view
  aggregation.
\newblock In {\em 2018 IEEE/RSJ International Conference on Intelligent Robots
  and Systems (IROS)},  1--8.
\newblock IEEE.

\bibitem[\protect\citeauthoryear{Ku, Pon, and
  Waslander}{2019}]{ku2019monocular}
Ku, J.; Pon, A.~D.; and Waslander, S.~L.
\newblock 2019.
\newblock Monocular 3d object detection leveraging accurate proposals and shape
  reconstruction.
\newblock In {\em Proceedings of the IEEE Conference on Computer Vision and
  Pattern Recognition},  11867--11876.

\bibitem[\protect\citeauthoryear{Lang \bgroup et al\mbox.\egroup
  }{2019}]{lang2019pointpillars}
Lang, A.~H.; Vora, S.; Caesar, H.; Zhou, L.; Yang, J.; and Beijbom, O.
\newblock 2019.
\newblock Pointpillars: Fast encoders for object detection from point clouds.
\newblock In {\em Proceedings of the IEEE Conference on Computer Vision and
  Pattern Recognition},  12697--12705.

\bibitem[\protect\citeauthoryear{Li \bgroup et al\mbox.\egroup
  }{2019a}]{li2019gs3d}
Li, B.; Ouyang, W.; Sheng, L.; Zeng, X.; and Wang, X.
\newblock 2019a.
\newblock Gs3d: An efficient 3d object detection framework for autonomous
  driving.
\newblock In {\em Proceedings of the IEEE Conference on Computer Vision and
  Pattern Recognition},  1019--1028.

\bibitem[\protect\citeauthoryear{Li \bgroup et al\mbox.\egroup
  }{2019b}]{DBLP:journals/corr/abs-1901-08373}
Li, X.; Guivant, J.~E.; Kwok, N.; and Xu, Y.
\newblock 2019b.
\newblock 3d backbone network for 3d object detection.
\newblock {\em CoRR} abs/1901.08373.

\bibitem[\protect\citeauthoryear{Liang \bgroup et al\mbox.\egroup
  }{2018}]{liang2018deep}
Liang, M.; Yang, B.; Wang, S.; and Urtasun, R.
\newblock 2018.
\newblock Deep continuous fusion for multi-sensor 3d object detection.
\newblock In {\em Proceedings of the European Conference on Computer Vision
  (ECCV)},  641--656.

\bibitem[\protect\citeauthoryear{Liang \bgroup et al\mbox.\egroup
  }{2019}]{liang2019multi}
Liang, M.; Yang, B.; Chen, Y.; Hu, R.; and Urtasun, R.
\newblock 2019.
\newblock Multi-task multi-sensor fusion for 3d object detection.
\newblock In {\em Proceedings of the IEEE Conference on Computer Vision and
  Pattern Recognition},  7345--7353.

\bibitem[\protect\citeauthoryear{Lin \bgroup et al\mbox.\egroup
  }{2017}]{lin2017focal}
Lin, T.-Y.; Goyal, P.; Girshick, R.; He, K.; and Doll{\'a}r, P.
\newblock 2017.
\newblock Focal loss for dense object detection.
\newblock In {\em Proceedings of the IEEE international conference on computer
  vision},  2980--2988.

\bibitem[\protect\citeauthoryear{Mousavian \bgroup et al\mbox.\egroup
  }{2017}]{mousavian20173d}
Mousavian, A.; Anguelov, D.; Flynn, J.; and Kosecka, J.
\newblock 2017.
\newblock 3d bounding box estimation using deep learning and geometry.
\newblock In {\em Proceedings of the IEEE Conference on Computer Vision and
  Pattern Recognition},  7074--7082.

\bibitem[\protect\citeauthoryear{Qi \bgroup et al\mbox.\egroup
  }{2017a}]{qi2017pointnet}
Qi, C.~R.; Su, H.; Mo, K.; and Guibas, L.~J.
\newblock 2017a.
\newblock Pointnet: Deep learning on point sets for 3d classification and
  segmentation.
\newblock In {\em Proceedings of the IEEE Conference on Computer Vision and
  Pattern Recognition},  652--660.

\bibitem[\protect\citeauthoryear{Qi \bgroup et al\mbox.\egroup
  }{2017b}]{qi2017pointnet++}
Qi, C.~R.; Yi, L.; Su, H.; and Guibas, L.~J.
\newblock 2017b.
\newblock Pointnet++: Deep hierarchical feature learning on point sets in a
  metric space.
\newblock In {\em Advances in neural information processing systems},
  5099--5108.

\bibitem[\protect\citeauthoryear{Qi \bgroup et al\mbox.\egroup
  }{2018}]{qi2018frustum}
Qi, C.~R.; Liu, W.; Wu, C.; Su, H.; and Guibas, L.~J.
\newblock 2018.
\newblock Frustum pointnets for 3d object detection from rgb-d data.
\newblock In {\em Proceedings of the IEEE Conference on Computer Vision and
  Pattern Recognition},  918--927.

\bibitem[\protect\citeauthoryear{Ronneberger, Fischer, and
  Brox}{2015}]{ronneberger2015u}
Ronneberger, O.; Fischer, P.; and Brox, T.
\newblock 2015.
\newblock U-net: Convolutional networks for biomedical image segmentation.
\newblock In {\em International Conference on Medical image computing and
  computer-assisted intervention},  234--241.
\newblock Springer.

\bibitem[\protect\citeauthoryear{Shi, Wang, and Li}{2019}]{shi2019pointrcnn}
Shi, S.; Wang, X.; and Li, H.
\newblock 2019.
\newblock Pointrcnn: 3d object proposal generation and detection from point
  cloud.
\newblock In {\em Proceedings of the IEEE Conference on Computer Vision and
  Pattern Recognition},  770--779.

\bibitem[\protect\citeauthoryear{Wang, An, and Cao}{2019}]{wang2019voxel}
Wang, B.; An, J.; and Cao, J.
\newblock 2019.
\newblock Voxel-fpn: multi-scale voxel feature aggregation in 3d object
  detection from point clouds.
\newblock {\em arXiv preprint arXiv:1907.05286}.

\bibitem[\protect\citeauthoryear{Wang \bgroup et al\mbox.\egroup
  }{2018}]{Wang_2018_CVPR}
Wang, S.; Suo, S.; Ma, W.-C.; Pokrovsky, A.; and Urtasun, R.
\newblock 2018.
\newblock Deep parametric continuous convolutional neural networks.
\newblock In {\em The IEEE Conference on Computer Vision and Pattern
  Recognition (CVPR)}.

\bibitem[\protect\citeauthoryear{Wang \bgroup et al\mbox.\egroup
  }{2019}]{wangcvpr2019}
Wang, Y.; Chao, W.-L.; Garg, D.; Hariharan, B.; Campbell, M.; and Weinberger,
  K.
\newblock 2019.
\newblock Pseudo-lidar from visual depth estimation: Bridging the gap in 3d
  object detection for autonomous driving.
\newblock In {\em CVPR}.

\bibitem[\protect\citeauthoryear{Yan, Mao, and Li}{2018}]{yan2018second}
Yan, Y.; Mao, Y.; and Li, B.
\newblock 2018.
\newblock Second: Sparsely embedded convolutional detection.
\newblock {\em Sensors} 18(10):3337.

\bibitem[\protect\citeauthoryear{Yang, Luo, and Urtasun}{2018}]{yang2018pixor}
Yang, B.; Luo, W.; and Urtasun, R.
\newblock 2018.
\newblock Pixor: Real-time 3d object detection from point clouds.
\newblock In {\em Proceedings of the IEEE conference on Computer Vision and
  Pattern Recognition},  7652--7660.

\bibitem[\protect\citeauthoryear{Zhou and Tuzel}{2018}]{zhou2018voxelnet}
Zhou, Y., and Tuzel, O.
\newblock 2018.
\newblock Voxelnet: End-to-end learning for point cloud based 3d object
  detection.
\newblock In {\em Proceedings of the IEEE Conference on Computer Vision and
  Pattern Recognition},  4490--4499.

\end{thebibliography}
\end{document}